\documentclass[10pt,twocolumn,letterpaper]{article}

\usepackage{cvpr}              
\usepackage{multirow}
\definecolor{cvprblue}{rgb}{0.21,0.49,0.74}
\usepackage[pagebackref,breaklinks,colorlinks,allcolors=cvprblue]{hyperref}
\usepackage{marvosym}
\def\paperID{18} 
\def\confName{CVPR}
\def\confYear{2026}

\title{Neural Surface Reconstruction from Sparse Views Using Epipolar Geometry}

\author{Xinhai Chang\\
Yuanpei College, Peking University\\
{\tt\small changxinhai@stu.pku.edu.cn}
\and
Kaichen Zhou\,{\Letter}\\
MIT Media Lab and EECS, MIT\\
{\tt\small zhouk777@mit.edu}
}

\begin{document}
\maketitle
\begin{abstract}
Reconstructing accurate surfaces from sparse multi-view images remains challenging due to severe geometric ambiguity and occlusions. Existing generalizable neural surface reconstruction methods primarily rely on cost volumes that summarize multi-view features using simple statistics (e.g., mean and variance), which discard critical view-dependent geometric structure and often lead to over-smoothed reconstructions.
We propose EpiS, a generalizable neural surface reconstruction framework that explicitly leverages epipolar geometry for sparse-view inputs. Instead of directly regressing geometry from cost-volume statistics, EpiS uses coarse cost-volume features to guide the aggregation of fine-grained epipolar features sampled along corresponding epipolar lines across source views. An epipolar transformer fuses multi-view information, followed by ray-wise aggregation to produce SDF-aware features for surface estimation.
To further mitigate information loss under sparse views, we introduce a geometry regularization strategy that leverages a pretrained monocular depth model through scale-invariant global and local constraints. Extensive experiments on DTU and BlendedMVS demonstrate that EpiS significantly outperforms state-of-the-art generalizable surface reconstruction methods under sparse-view settings, while maintaining strong generalization without per-scene optimization.
\end{abstract}    
\section{Introduction}
Surface reconstruction from multi-view images is a fundamental problem in computer vision~\cite{yao2018mvsnet,gu2020cascade,ding2022transmvsnet}, robotics~\cite{zhou2022devnet}, and virtual reality~\cite{middelberg2014scalable,zhou2021vmloc}. 
Traditional multi-view stereo pipelines decompose the problem into depth estimation, depth fusion, and surface meshing, which often leads to cumulative errors and brittle performance under challenging conditions.
Recent advances in neural implicit representations have significantly reshaped this landscape, enabling end-to-end surface reconstruction and novel view synthesis.
Methods such as NeuS~\cite{wang2021neus} and NeuS2~\cite{wang2023neus2}, inspired by NeRF~\cite{mildenhall2021nerf}, represent geometry using Signed Distance Functions (SDFs) and volume rendering, achieving high-quality reconstructions when dense views or strong supervision are available.

However, reconstructing accurate surfaces from \emph{sparse} multi-view inputs remains highly challenging.
Limited viewpoints introduce severe geometric ambiguity and occlusions, while the lack of dense supervision restricts the quality of learned geometry.
Existing approaches attempt to alleviate this issue by incorporating additional priors, such as sparse Structure-from-Motion (SfM) point clouds~\cite{schonberger2016structure} or depth maps.
Although effective, these methods typically rely on costly per-scene optimization and struggle to generalize across scenes.
More recent generalizable methods, such as SparseNeuS~\cite{long2022sparseneuss}, avoid per-scene training but suffer from resolution limitations imposed by memory constraints, often producing over-smoothed surfaces.
VolRecon~\cite{ren2023volrecon} improves reconstruction fidelity by supervising with high-resolution ground-truth depth maps, yet this requirement substantially limits its applicability in real-world scenarios.

\begin{figure}[tb]
    \centering
  \includegraphics[width=0.5\textwidth]{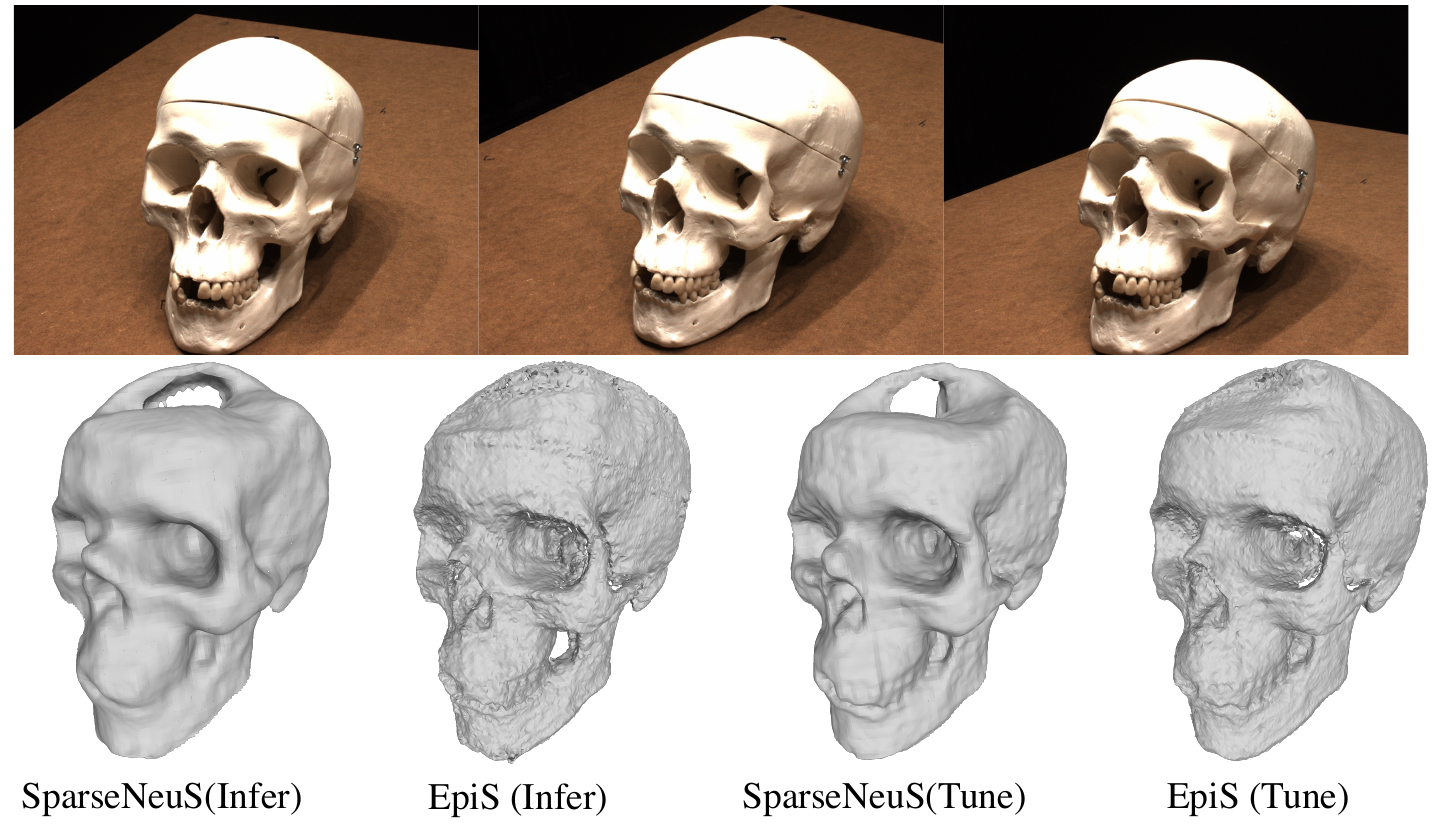}
    \vspace{-0.6cm}
    \caption{\textbf{Reconstruction results on the DTU dataset.} Our approach has remarkable generalization capabilities across various scenes, successfully reconstructing neural surfaces using only three source images through fast network inference. Notably, the reconstruction quality of our fast inference process surpasses that of SparseNeuS, offering enhanced accuracy and fidelity. Additionally, our results can be further refined through per-scene adjustments. (All meshes are visualized with the help of MeshLab2022).}
    \label{fig:fig1}
\vspace{-0.5cm}
\end{figure}

A key limitation shared by existing generalizable sparse-view methods lies in their reliance on cost volumes that summarize multi-view features using simple statistics, such as mean and variance.
While effective for capturing coarse geometric cues, these statistics discard view-dependent correspondences encoded along epipolar lines, which are critical for resolving fine-grained geometry under sparse views.
As a result, important structural details are lost during feature aggregation, leading to blurred or incomplete surface reconstructions.

To address this limitation, we propose \textbf{EpiS}, a fast and generalizable neural surface reconstruction framework tailored for sparse-view inputs.
Instead of directly regressing geometry from cost-volume statistics, EpiS uses coarse cost-volume features to \emph{guide the aggregation of fine-grained epipolar features} sampled from multiple source views.
An epipolar-guided attention mechanism explicitly fuses view-dependent information along corresponding epipolar lines, while a subsequent ray-wise aggregation module integrates features along each target ray to produce SDF-aware representations.
This design preserves critical geometric structure while remaining compatible with generalizable training.

To further compensate for missing information under sparse views, we leverage a pretrained monocular depth model as a geometry regularizer.
Rather than enforcing absolute depth consistency, we introduce scale-invariant global and local constraints through a triplet-based global regularization and a gradient-based local regularization, improving geometric stability without requiring ground-truth depth supervision.

Our contributions are summarized as follows:
\begin{itemize}
    \item We propose \textbf{EpiS}, a generalizable neural surface reconstruction framework that explicitly incorporates epipolar geometry to preserve fine-grained multi-view structure under sparse-view inputs.
    \item We introduce an epipolar-guided feature aggregation strategy that uses cost-volume information to steer multi-view epipolar fusion, followed by ray-wise aggregation for SDF prediction.
    \item We design scale-invariant depth regularization strategies leveraging a pretrained monocular depth model, improving reconstruction accuracy under sparse views without per-scene optimization.
    \item Extensive experiments demonstrate that EpiS consistently outperforms state-of-the-art generalizable surface reconstruction methods, particularly in sparse-view and cross-scene settings (Fig.~\ref{fig:fig1}).
\end{itemize}
\section{Related Works}
\subsection{Multi-view depth estimation}
Reconstructing 3D geometry from multi-view images poses a fundamental challenge in 3D vision, traditionally approached via depth-based or voxel-based methodologies. 
Multi-view stereo (MVS) methods, such as those by ~\cite{campbell2008using,stereopsis2010accurate,ji2017surfacenet}, rely on stereo correspondence for reconstructing depth maps. 
While early MVS methods utilized hand-crafted similarity metrics, recent advancements integrate deep learning for more precise matching. These approaches span volumetric~\cite{ji2020surfacenet+,kar2017learning,kutulakos2000theory}, point cloud-based~\cite{gropp2020implicit,lhuillier2005quasi}, and depth map-based methodologies~\cite{schonberger2016pixelwise,tola2012efficient,yao2018mvsnet,aanaes2016large,gu2020cascade}. 
Volumetric and point cloud-based techniques directly model objects but face memory limitations~\cite{campbell2008using,stereopsis2010accurate}. 
In contrast, depth map-based methods offer flexibility by separating depth map estimation and fusion, yielding commendable performance on diverse benchmarks~\cite{ji2017surfacenet}. 
However, their intricate processes, including depth filtering and fusion, may introduce cumulative errors. 
Despite recent strides in neural implicit representation, their performance trails behind state-of-the-art MVS methods~\cite{wang2024dust3r,wang2025vggt,zhou2026page}. 
Addressing this gap, our paper introduces EpiS, showcasing superior performance compared to MVSNet~\cite{yao2018mvsnet} and COLMAP~\cite{schonberger2016structure} under sparse view conditions through neural implicit representation.

\subsection{Neural Implicit Surface Estimation}
Neural implicit functions have recently gained traction as effective representations of 3D geometry~\cite{genova2019learning,mescheder2019occupancy,michalkiewicz2019implicit,niemeyer2019occupancy,park2019deepsdf,peng2020convolutional} and appearance~\cite{liu2020neural,liu2020dist,muller2022instant,oechsle2019texture,pumarola2021d,sun2022direct}. 
These functions are utilized in both surface and volume rendering approaches to realize 3D geometry reconstruction without the need for 3D supervision. 
While surface rendering~\cite{niemeyer2020differentiable,yariv2020multiview,zhang2021learning} methods focus on single surface intersection points, volume rendering~\cite{oechsle2021unisurf,wang2021neus,yariv2021volume,yu2022monosdf} considers multiple points along the ray, resulting in more impressive results. 
However, both approaches require expensive per-scene optimization and struggle to generalize to new scenes.
Successful attempts at generalization~\cite{johari2022geonerf,chen2021mvsnerf,wang2021ibrnet,yu2021pixelnerf}, in novel view synthesis based on differentiable rendering have leveraged sparse views and radiance information.
While these approaches yield more complete surfaces compared to traditional methods, they may struggle with complex structures and sharp corners.

In response to these concerns, recent research has aimed to enhance generalizability and accommodate sparse input by integrating traditional MVS techniques with differentiable rendering.
SparseNeuS~\cite{long2022sparseneuss}, for example, reconstructs surfaces from nearby viewpoints but may lack detail and suffer from error accumulation. Furthermore, its two-stage training process prolongs training times and complexity.
In contrast, VolRecon~\cite{ren2023volrecon} and ReTR~\cite{liang2024retr} rely on detailed depth ground truth for training, limiting their applicability and fine-tuning stability. Despite attempts at fine-tuning, their results remain inconsistent.
Another recent approach, GenS~\cite{peng2024gens}, requires more views than the conventional sparse view setting to achieve comparable results.
\begin{figure*}[tb]
    \centering
  \includegraphics[width=0.9\textwidth]{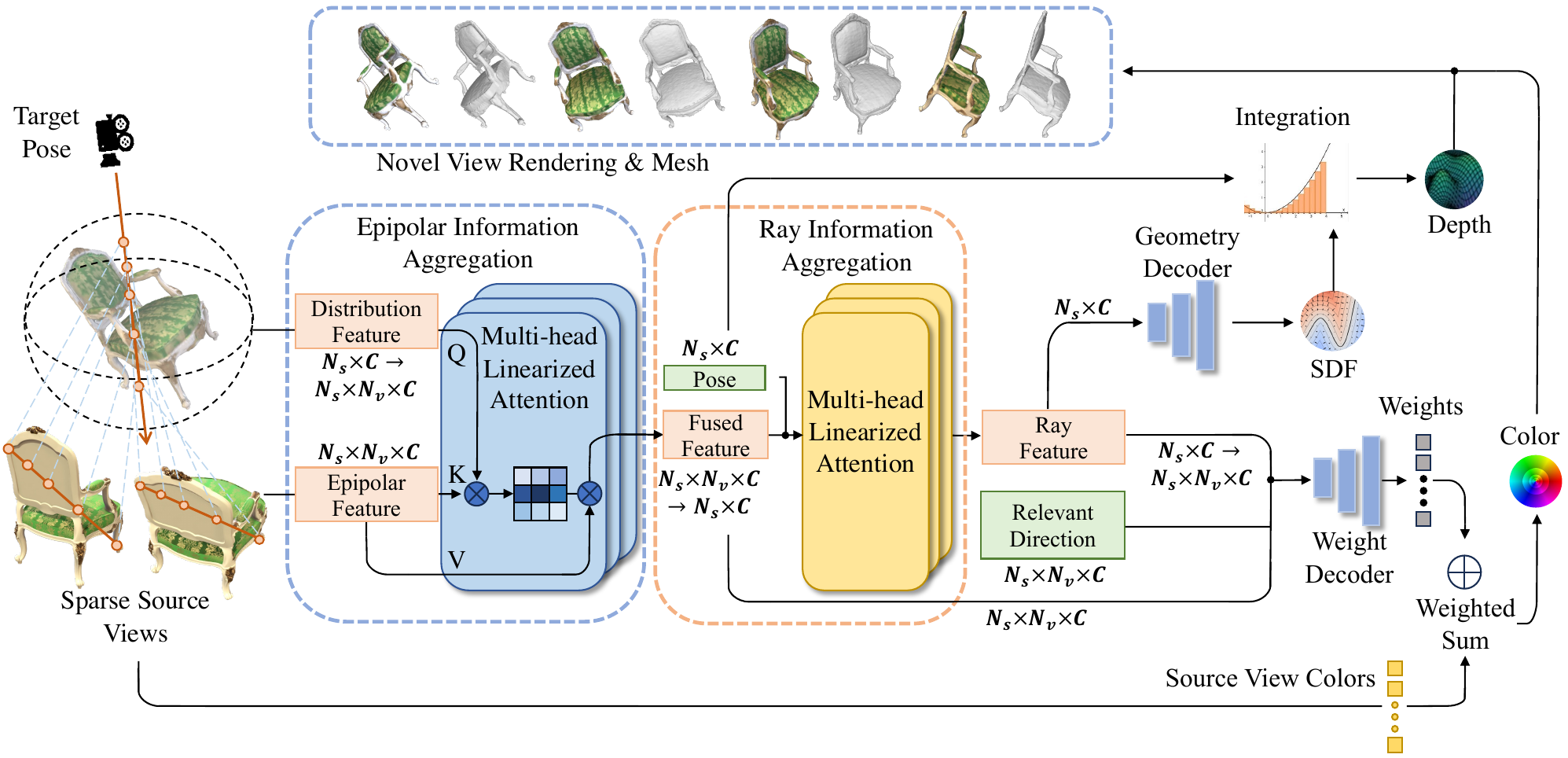}
  \vspace{-0.6cm}
    \caption{\textbf{Illustration of the Pipeline.} Given a ray in the target view, it is projected onto source views to extract the epipolar feature and distribution feature (variance and mean) using a cost volume. Subsequently, the distribution features are utilized as queries, while the epipolar features serve as keys and values for cross-attention transformers, facilitating cross-view epipolar feature fusion. This fused feature set serves as input for subsequent ray transformers, enabling feature aggregation along the target ray. Finally, the resulting feature is used in the geometry MLP and weight decoder to predict corresponding signed distance functions (SDF) and multi-view color weights.}
    \label{fig:fig2}
    \vspace{-0.6cm}
\end{figure*}

This paper introduces EpiS, a method capable of achieving remarkable 3D reconstruction under conventional sparse view settings without requiring 3D ground truth supervision. Additionally, we propose efficient regularization techniques to ensure optimal fine-tuning performance using sparse view inputs as shown in Figure.~\ref{fig:fig1}.
\section{Methodology}
Given a set of sparse input images $\{I_i\}_{i=1}^{n}$ and their corresponding camera poses $\{P_i\}_{i=1}^{n}$, our goal is to reconstruct accurate surface geometry while generalizing across scenes without per-scene optimization, and
to support stable fine-tuning under sparse supervision.
\textbf{(1)}
Most existing sparse-view surface reconstruction methods, such as SparseNeuS~\cite{long2022sparseneuss} and GenS~\cite{peng2024gens}, rely on cost volumes that summarize multi-view information using simple statistics (e.g.,
mean and variance).
While effective for coarse geometry, this representation discards fine-grained view correspondences along epipolar lines, often leading to over-smoothed or ambiguous surfaces under sparse views.
To overcome this limitation, we combine coarse cost-volume cues with explicit epipolar feature aggregation across source views.
An epipolar transformer is introduced to fuse multi-view information along epipolar lines, enabling more precise and geometry-aware surface estimation.
\textbf{(2)}
Fine-tuning in sparse-view settings is further challenged by geometric ambiguity and occlusions.
To stabilize optimization, we leverage pretrained monocular depth models as a source of geometric prior.
Instead of relying on absolute depth supervision, we introduce two lightweight regularization terms that exploit relative depth consistency, improving depth and surface estimation accuracy during fine-tuning without requiring metric ground-truth depth.

\subsection{Preliminaries}
The goal of EpiS is to produce appearance and geometry information for a given novel view, following the pipeline established by NeuS~\cite{wang2021neus}.

\textbf{Depth Information Estimation}: This is realized by a network that takes a feature considering the 3D point position as input and predicts the surface information.
The surface is represented by the zero-level set of the Signed Distance Function (SDF): $S = \{ p \in \mathbb{R}^3 | \text{sdf}_{\theta}(p) = 0 \}.$
Following NeuS~\cite{wang2021neus}, we compute opaque density function $\rho(t)$ from $S$. 
To make the depth supervision differential, we could derive the rendered depth as in~\cite{zhou2022devnet}:
\begin{equation} \label{eq::depthrender}
    \hat{D}=\sum_{j=1}^{N_S} T_j \alpha_j t_j, \quad T_j=\prod_{k=1}^{j-1}\left(1-\alpha_k\right),
\end{equation}
where $\alpha_j = 1- \exp (-\int_{t_j}^{t_{j+1}} \rho(t) dt)$; $t$ is the z value from the sampled point to the camera origin and $N_S$ is number of sampled points along each ray.

\textbf{Appearance Information Estimation}: Following IBRNET~\cite{wang2021ibrnet}, we make use of the color blending technique. Give a point $p_t$ on the ray of a target frame, this point is projected into source frames to get the color information $c_s$ of multiple pixels. A decoder would be used to predict the weights $w_s$ of each source frame. 
Based on the volume rendering in NeRF~\cite{mildenhall2021nerf}, the final color is computed as:
\begin{equation} \label{eq::colorrender}
    \hat{C}=\sum_{j=1}^{N_S} T_j \alpha_j c_j, \quad c_j = \sum_{s=1}^{N_V} w_s * c_s,
\end{equation}
where $c_j$ is the predicted radiance and $N_V$ is the number of source views.

\subsection{Generalizable Framework}
\subsubsection{Cost Volume Construction}
To construct a cost volume for the target camera pose with $N_V$ source frames, we first extract 2D feature maps $\{\boldsymbol{F_i}\}_{i=0}^{N_V}$ from the input images $\{I_i\}_{i=0}^{N_V}$. 
Then, we construct a corresponding bounding box $B$ for this target pose. To get the feature for each grid $b$ in this bounding box. We project each grid on each source plane and get its feature $\{\boldsymbol{F_i}(P_i(b))\}_{i=0}^{N_V}$, where $P_i$ is the transformation matrix between the target pose and the source pose. 
Following previous methods \cite{yao2018mvsnet}, we calculate the variance and mean of the projected features for each grid $b$ of the bounding box to create a cost volume.
\begin{equation}
\quad B_1(b) = \operatorname{Var}\left({{\boldsymbol{F_i}(P_i(b))}}_{i=1}^{N_V}\right), 
\end{equation}
\begin{equation}
B_2(b) = \operatorname{Mean}\left({{\boldsymbol{F_i}(P_i(b))}}_{i=1}^{N_V}\right),
\end{equation}
Then, we employ a sparse 3D convolutional neural network $\Psi$ to process the cost volume and get the coarse geometry aware feature volume $\hat{B}$.

\subsubsection{Epipolar \& Ray Information Aggregation}
Relying solely on variance and means to characterize the distribution of multi-view information overlooks the intricate relationships between multiple views. 
To address this limitation, we propose an approach that aggregates multi-view information along epipolar lines from multiple source views, while still considering distribution information.

Specifically, for each ray originating from a pixel, EpiS samples $N_S$ points along the ray and projects them onto each source view. 
This process yields epipolar features from each source view. 
Subsequently, EpiS utilizes features from the cost volume as the query and the epipolar features from each source view as the key and value within a linearized attention mechanism~\cite{katharopoulos2020transformers}. 
This mechanism facilitates the fusion of multi-view epipolar information. Finally, a linear attention mechanism is employed to fuse information along the target ray.

\textbf{(a) Epipolar Aggregation}: 
This aggregation is realized through constructing a cross-attention transformer, as shown in Figure.~\ref{fig:fig2}.
Given a ray $r_t = \{p_i\}_{i=1}^{i=N_S}$ from the sampled pixel of the target frame, EpiS extracts the corresponding feature $\boldsymbol{F_B} \in \mathbb{R}^{N_S \times C}$ from cost-volume $\hat{B}$ and the multi-view epipolar feature $\boldsymbol{F_E} \in \mathbb{R}^{N_V\times N_S\times C}$ with the help of $\boldsymbol{F_i}$ and $P_i$. 
EpiS initially applies corresponding matrices to process them and compute the query, key, and value as follows:
\begin{equation}
\boldsymbol{Q}=\boldsymbol{F_B}\boldsymbol{W_{Q}},\,\boldsymbol{K}=\boldsymbol{F_E}\boldsymbol{W_{K}},\,\boldsymbol{V}=\boldsymbol{F_E}\boldsymbol{W_{V}},
\end{equation}
where $\boldsymbol{W_{Q}},\boldsymbol{W_{K}},\boldsymbol{W_{V}}\in \mathbb{R}^{C\times C}$.
Considering that the query, key, and value have different dimensions, EpiS repeats the query feature along the first dimension and we have $\boldsymbol{Q} \in \mathbb{R}^{N_S \times C} \rightarrow \boldsymbol{Q} \in \mathbb{R}^{N_V \times N_S \times C}$. To ensure Linearized Attention is applied across both multi-view and channel dimensions, we apply dimension permutation to the query, key, and value. In this way, we have $\boldsymbol{Q},\boldsymbol{K},\boldsymbol{V} \in \mathbb{R}^{N_S \times N_V \times C}$. These matrices are then split into $h$ heads $\boldsymbol{Q}=\{\boldsymbol{Q}^{i}\}_{i=1}^{h}$, $\boldsymbol{K}=\{\boldsymbol{K}^{i}\}_{i=1}^{h}$, and $\boldsymbol{V}=\{\boldsymbol{V}^{i}\}_{i=1}^{h}$, each with $d=C/h$ channels. The cost-volume information is used to guide the learning process through the following Linearized Attention mechanism as: 
\begin{equation}
    \boldsymbol{X}^{i}_r = \frac{\phi(\boldsymbol{Q}^{i}_r)^T (\sum \phi(\boldsymbol{K}^{i}_r) (\boldsymbol{V}^{i}_r)^T)}{\phi(\boldsymbol{Q}^{i}_r)^T \sum \phi(\boldsymbol{K}^{i}_r)},
\end{equation} 
where subscripting a matrix with $i$ returns the $i-th$ row along $N_V$ dimension and $\phi(.)$ is the kernal function. In our implementation, we use the same kernel function as in \cite{katharopoulos2020transformers}, which could be written as $\phi(.)=\text{elu}(x) + 1$. Ultimately, we obtain fused feature $\boldsymbol{X}=\{\boldsymbol{X}^{i}\}_{i=1}^{h} \in \mathbb{R}^{N_S \times N_V \times C}$.

\begin{figure}[tb]
    \centering
  \includegraphics[width=0.45\textwidth]{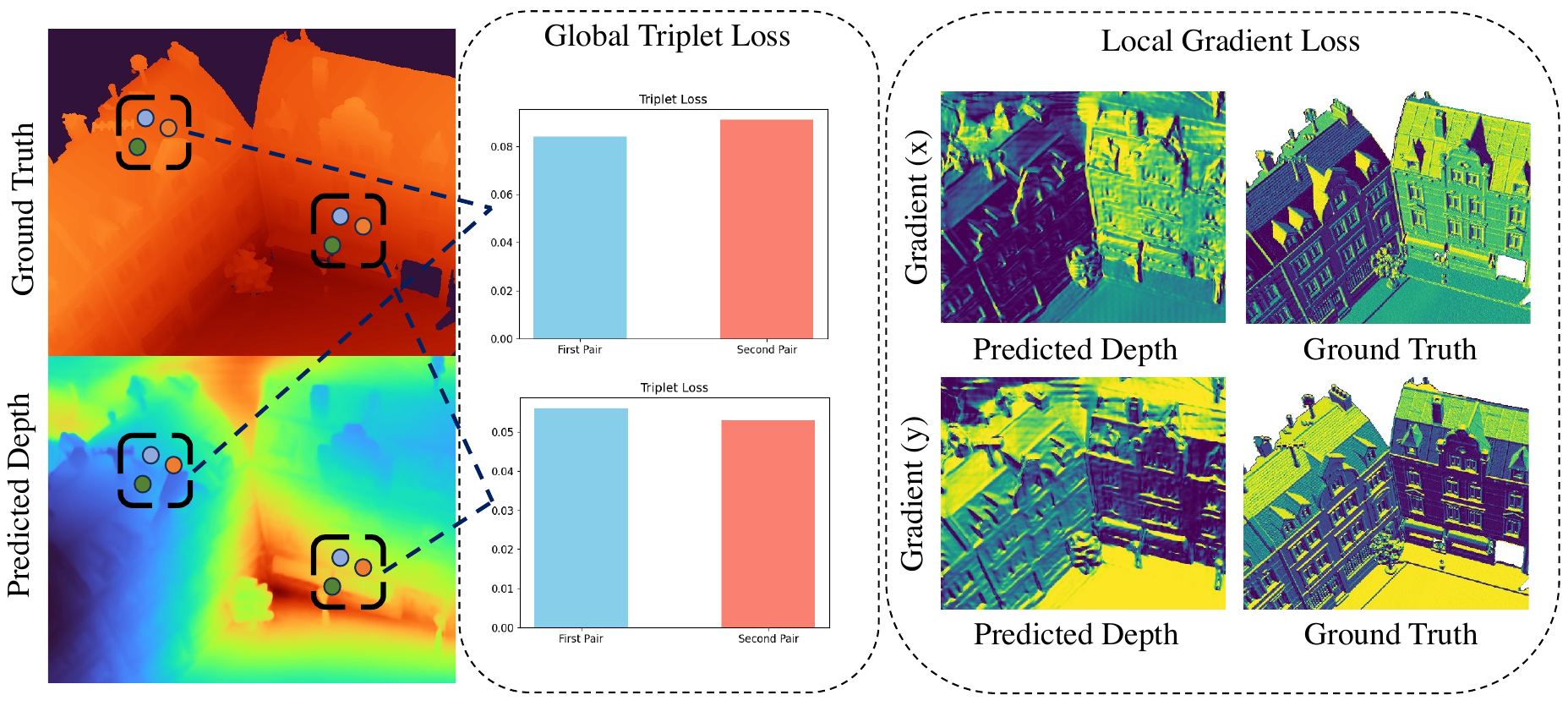}
    \caption{\textbf{Visualization of Our Fine-Tuning Strategy Designs.} On the left, we present the predicted and ground truth depth maps. In the middle, we illustrate the triplet loss. On the right, we showcase the derivative gradients along the X and Y axes of the images.}
    \label{fig:fig3}
    \vspace{-0.6cm}
\end{figure}

\textbf{(b) Ray Aggregation}: Efficient aggregation of epipolar information facilitates the integration of information from epipolar lines across various source views, as shown in Figure.~\ref{fig:fig2}.
However, taking into account the feature of SDF, which is zero at the surface points and increases or decreases when away from the surface. 
We also need to consider the information along the ray. 
To realize this task, we first take the mean of $\boldsymbol{X}$ along the $N_V$ dimension and we have $\boldsymbol{X}' \in \mathbb{R}^{N_S \times C}$. 

\begin{figure*}[tb]
    \centering
  \includegraphics[width=\textwidth]{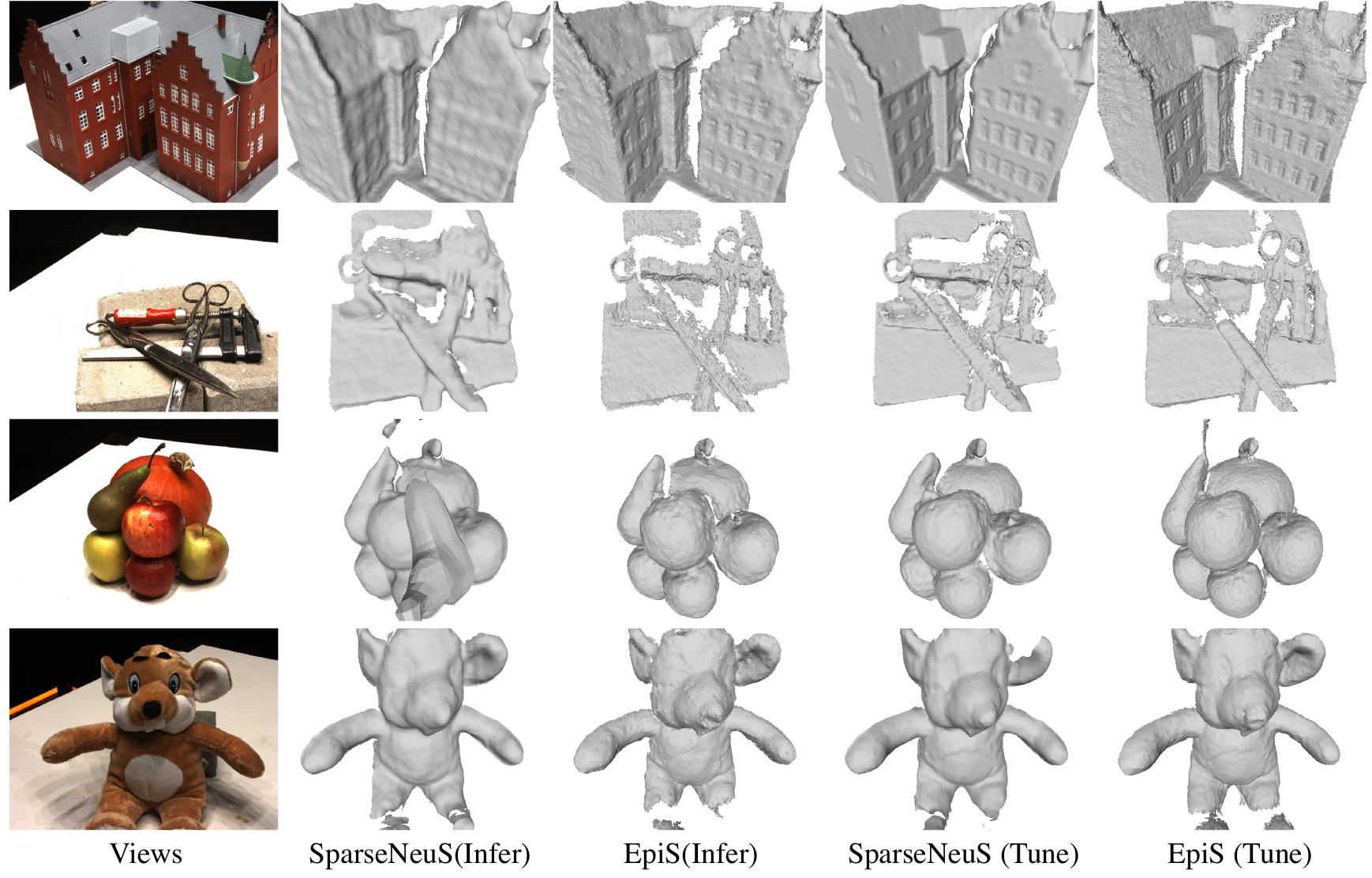}
    \caption{\textbf{Visualization results on the DTU dataset.} EpiS produces precise outcomes without requiring fine-tuning. Moreover, fine-tuning further enhances the realism of our results, which is evident in the comparison.}
    \label{fig:fig5}
    \vspace{-0.5cm}

\end{figure*}

Furthermore, given that $N_S$ samples are randomly selected along each ray, the distance between each position is uncertain, particularly during the fine-sampling stage, as observed in NeRF~\cite{mildenhall2021nerf}. 
To incorporate this information, we adopt the approach used in previous work, NeRF~\cite{mildenhall2021nerf} to embed 3D position information as following $emd(\boldsymbol{x})$. 
We first concatenate this embedding information with feature information as $\boldsymbol{\hat{X}} = concat(\boldsymbol{X}', emb(\boldsymbol{x}))$. Finally following the previous section, we first process this feature with the help of linear mappings: $\boldsymbol{\hat{Q}}=\boldsymbol{\hat{X}}\boldsymbol{\hat{W_{Q}}},\,\boldsymbol{\hat{K}}=\boldsymbol{\hat{X}}\boldsymbol{\hat{W_{K}}},\,\boldsymbol{\hat{V}}=\boldsymbol{\hat{X}}\boldsymbol{\hat{W_{V}}}$. The final ray feature could be written as $\boldsymbol{\hat{X}} \in \mathbb{R}^{N_S \times C}$.

\subsubsection{Geometry Decoder \& Weights Decoder}
To compute the final weights and Signed Distance Function (SDF), we employ a decoder, as shown in Figure.~\ref{fig:fig2}, similar to the approach used in NeuS~\cite{wang2021neus}. Specifically, the Geometry Decoder $f_{\theta}$ utilizes the final ray feature $\boldsymbol{\hat{X}}$ as input to predict the final SDF. The decision not to include pose information as an input is because $\boldsymbol{\hat{X}}$ inherently encodes pose information. Additionally, the Weight Decoder takes the ray feature $\boldsymbol{\hat{X}}$, fused feature $\boldsymbol{X}$, and relevant directional information as inputs to predict the final color weights.

\subsection{Fine-Tuning Strategy for Accuracy}

Pre-trained large-scale monocular depth estimation models, trained on extensive datasets, demonstrate efficiency in estimating depth across diverse scenarios~\cite{bhat2023zoedepth,ranftl2020towards}. However, these models typically output depth information without scaling. While they perform well in estimating relative depth, absolute depth estimation can be challenging. Often, the generated depth lacks a scale factor and offset compared to the ground truth, as illustrated in DynPoint~\cite{zhou2024dynpoint}, where $d_{gt} = \alpha \hat{d} + \beta$. This feature complicates the use of generated depths for supervising depth estimation tasks. To address this, we propose both a global triplet loss function and a local gradient loss function.

\textbf{(a) Global Triplet Loss}:
Addressing the challenge posed by the scale discrepancy between predicted depth and ground truth depth, directly employing predicted depth for supervision proves difficult. However, by capitalizing on the sampling of $NR$ from the target frame during each iteration for training, exploiting the relative relationship between different rays becomes viable. Specifically, given sampled rays $r_s$, two rays $r_1$ and $r_2$ are randomly selected from $NR$. Subsequently, the global triplet loss function can be expressed as:

\begin{equation}
\mathcal{L}_{\text {global}}  = \left( (\hat{d_1} - \hat{d_s}) \times (\tilde{d_2} - \tilde{d_s}) - (\hat{d_2} - \hat{d_s}) \times (\tilde{d_1} - \tilde{d_s})\right)^2,
\end{equation}
where $\hat{d_1}, \hat{d_2}, \hat{d_s}$ represent the estimated depth by EpiS, and $\tilde{d_1}, \tilde{d_2}, \tilde{d_s}$ represent the estimated depth by the Pretrained Depth Model. Please note that while this loss function is named "global," it is applied only to local patches of the depth map. This approach is adopted because ensuring the relevance of the relationship between two distant pixels of the depth map cannot be guaranteed. The intuition of this loss function is demonstrated in the left part of Figure.~\ref{fig:fig3}.

\textbf{(b) Local Gradient Loss}: 
The global triplet loss primarily considers the overall structure of the depth map, often overlooking detailed local information. To incorporate local gradients, we utilize the partial derivatives of the depth map along the x and y axes. The partial derivative vectors are defined as  $\boldsymbol{\hat{v}} = (\frac{\partial \hat{d}}{\partial x},  \frac{\partial \hat{d}}{\partial y})$ and $\boldsymbol{\Tilde{v}} = (\frac{\partial \Tilde{d}}{\partial x},  \frac{\partial \Tilde{d}}{\partial y})$. The loss function could be written as:
\begin{equation}
    \mathcal{L}_{\text {local}} = (1 - \frac{\boldsymbol{\hat{v}}\cdot\boldsymbol{\Tilde{v}}}{||\boldsymbol{\hat{v}}||\cdot||\boldsymbol{\Tilde{v}}||})^2.
\end{equation}
This loss function is employed to regularize the direction of the partial derivative vector while disregarding their absolute value. The intuition of this loss function is demonstrated in the right part of Figure.~\ref{fig:fig3}.
\begin{table*}[ht]
\caption{Quantitative results of sparse view reconstruction on 15 testing scenes from the DTU dataset. The upper part presents the performance of generalizable models, while the lower part displays the performance of fine-tuning models. If the result of EpiS outperforms all other methods, it will be \textbf{highlighted}. If the result of EpiS is the second best among all methods, it will be \underline{underlined}.}
\vspace{-0.3cm}
\resizebox{\textwidth}{!}{\renewcommand{\arraystretch}{1.1}
\begin{tabular}{lcccccccccccccccc} \hline
Method & 24   & 37   & 40   & 55   & 63   & 65   & 69   & 83   & 97   & 105  & 106  & 110  & 114  & 118  & 122  & Mean$ \downarrow$  \\ \hline
PixelNerf~\cite{yu2021pixelnerf} & 5.13 & 8.07 & 5.85 & 4.40 & 7.11 & 4.64 & 5.68 & 6.76 & 9.05 & 6.11 & 3.95 & 5.92 & 6.26 & 6.89 & 6.93 & 6.28 \\
IBRNet~\cite{wang2021ibrnet} & 2.29 & 3.70 & 2.66 & 1.83 & 3.02 & 2.83 & 1.77 & 2.28 & 2.73 & 1.96 & 1.87 & 2.13 & 1.58 & 2.05 & 2.09 & 2.32 \\
MVSNerf~\cite{chen2021mvsnerf} & 1.96 & 3.27 & 2.54 & 1.93 & 2.57 & 2.71 & 1.82 & 1.72 & 2.29 & 1.75 & 1.72 & 1.47 & 1.29 & 2.09 & 2.26 & 2.09 \\
SparseNeuS~\cite{long2022sparseneuss} & 1.68 & 3.06 & 2.25 & 1.10 & 2.37 & 2.18 & 1.28 & 1.47 & 1.80 & 1.23 & 1.19 & 1.17 & 0.75 & 1.56 & 1.55 & 1.64 \\
\textbf{EpiS} & \textbf{1.11} & \textbf{2.71} & \textbf{1.85} & \underline{1.12} & \textbf{1.47} & \textbf{1.69} & \underline{1.05} & \textbf{1.45} & \textbf{1.35} & \textbf{0.97} & \underline{1.21} & \underline{1.35} & \textbf{0.71} & \textbf{1.20} & \textbf{1.22} & \textbf{1.36}\\
\hline 
UniSurf~\cite{oechsle2021unisurf} & 5.08 & 7.18 & 3.96 & 5.30 & 4.61 & 2.24 & 3.94 & 3.14 & 5.63 & 3.40 & 5.09 & 6.38 & 2.98 & 4.05 & 2.81 & 4.39 \\
NeuS~\cite{wang2021neus} & 4.57 & 4.49 & 3.97 & 4.32 & 4.63 & 1.95 & 4.68 & 3.83 & 4.15 & 2.50 & 1.52 & 6.47 & 1.26 & 5.57 & 6.11 & 4.00 \\
VolSDF~\cite{yariv2021volume} & 4.03 & 4.21 & 6.12 & 0.91 & 8.24 & 1.73 & 2.74 & 1.82 & 5.14 & 3.09 & 2.08 & 4.81 & 0.60 & 3.51 & 2.18 & 3.41 \\
IBRNet (ft)~\cite{wang2021ibrnet} & 1.67 & 2.97 & 2.26 & 1.56 & 2.52 & 2.30 & 1.50 & 2.05 & 2.02 & 1.73 & 1.66 & 1.63 & 1.17 & 1.84 & 1.61 & 1.90 \\
Colmap~\cite{schonberger2016structure} & 0.90 & 2.89 & 1.63 & 1.08 & 2.18 & 1.94 & 1.61 & 1.30 & 2.34 & 1.28 & 1.10 & 1.42 & 0.76 & 1.17 & 1.14 & 1.52 \\
SparseNeuS (ft)~\cite{long2022sparseneuss} & 1.29 & 2.27 & 1.57 & 0.88 & 1.61 & 1.86 & 1.06 & 1.27 & 1.42 & 1.07 & 0.99 & 0.87 & 0.54 & 1.15 & 1.18 & 1.27 \\
MVSNet~\cite{yao2018mvsnet} & 1.05 & 2.52 & 1.71 & 1.04 & 1.45 & 1.52 & 0.88 & 1.29 & 1.38 & 1.05 & 0.91 & 0.66 & 0.61 & 1.08 & 1.16 & 1.22 \\
\textbf{EpiS (ft)} & \underline{0.93} & \textbf{2.13} & \textbf{1.32} & \textbf{0.87} & \textbf{1.01} & \textbf{1.56} & \textbf{0.84} & \textbf{1.21} & \textbf{1.10} & \textbf{0.84} & \textbf{0.79} & \underline{0.89} & \textbf{0.51} & \textbf{1.05} & \textbf{1.06} & \textbf{1.07} \\ \hline 
\end{tabular}
}
\vspace{-0.3cm}
\label{table:dtusparse}
\end{table*}

\subsection{Loss Function}
Given that EpiS relies solely on color information for supervising neural surface reconstruction, akin to prior neural surface reconstruction methods~\cite{wang2023neus2}, we utilize the following loss functions for training:
\begin{equation}
\mathcal{L}= \mathcal{L}_{\text {color}}+\lambda_1 \mathcal{L}_{\text {eik}} + \lambda_2 \mathcal{L}_{\text {sparse}} + \lambda_3 \mathcal{L}_{\text {global}} + \lambda_4 \mathcal{L}_{\text {local}} 
\end{equation}

The first four loss functions are frequently employed in sparse view neural surface reconstruction tasks. The color loss function~\cite{wang2023neus2} is used to compute the distance between the predicted color from Eqn.~\ref{eq::colorrender}, which could be written as:
\begin{equation}
    \mathcal{L}_{color}= \frac{1}{\left\| \text{Patch} \right\|} \sum_{ pix\in \text{Patch}} | \hat{c}(pix) - c(pix)|,
\end{equation}
where $\text{Patch}$ means the sampled patch used in one iteration and $\text{pix}$ represents pixels within this patch.

The second Eikonal loss function~\cite{wang2023neus2} is used to regularize predicted SDF value predicted by our network, which could be written as:
\begin{equation}
    \mathcal{L}_{eik}=\frac{1}{\left\| \mathbb{P} \right\|} \sum_{p \in \mathbb{P}} (\left\| \nabla\text{sdf}_{\theta}(p) \right\|_2 -1)^2
\end{equation}
where $\mathbb{P}$ is the set of all sampled points based on the sampled patch. This loss function forces network $\text{sdf}_{\theta}(p)$ to have a unit gradient. 

To oversee the points beneath the predicted surface and prevent the occurrence of free surfaces, akin to \cite{long2022sparseneuss}, we integrate the sparse loss function term. This term aims to minimize the distance of points beneath the surface to zero as much as possible:
\begin{equation}
\label{sparse_lossterm}
    \mathcal{L}_{sparse}=\frac{1}{\left\|\mathbb{P} \right\| } \sum_{p \in \mathbb{P}} \exp \left(-\tau \cdot \left|\text{sdf}_{\theta}(p) \right|\right),
\end{equation}
where $\tau$ is the hyperparameter.

\section{Experiments}

\subsection{Experimental Settings}
\subsubsection{Baselines}

In our study, we follow the methodology outlined by MVSNet~\cite{chen2021mvsnerf} for dense view analysis, utilizing the first 49 images for training. Our method undergoes comparison with state-of-the-art approaches from three classes: generic neural rendering methods, including PixelNerf~\cite{yu2021pixelnerf}, IBRNet~\cite{wang2021ibrnet}, MVSNerf~\cite{chen2021mvsnerf} and SparseView~\cite{long2022sparseneuss}; per-scene optimization-based neural surfaces reconstruction methods like IDR~\cite{yariv2020multiview}, NeuS~\cite{wang2023neus2}, VolSDF~\cite{yariv2021volume}, and UniSurf~\cite{oechsle2021unisurf}; and the classic MVS method COLMAP~\cite{schonberger2016structure} and MVSNet~\cite{yao2018mvsnet}. All methods use three images as input. It's worth noting that MVS methods, unlike neural implicit reconstruction, do not explicitly model scene parameters and cannot render novel views.

Besides, It's worth noting that while GenS shows promising results, conducting a direct comparison poses challenges. 
The previous study~\cite{long2022sparseneuss} focused on utilizing three views (N = 3), while GenS~\cite{peng2024gens} employs four views (N = 4), leading to significant differences in results. 
Moreover, the specific configuration of the four views used in GenS remains undisclosed, hindering clarity on their impact. Additionally, GenS's use of 19 views for their Dense View experiment further complicates comparisons, especially considering the undisclosed rationale behind this choice. 
Due to these factors, we refrain from providing a direct comparison between EpiS and GenS.

\begin{table}[tb]
\caption{Depth evaluation on the DTU dataset. The result of mean absolute error (Abs.) is in millimeters. The results of threshold percentage ($<1mm$, $<2mm$, $<4mm$) and mean absolute relative error (Rel.) are in percentage (\%). If the result of EpiS outperforms all other methods, it will be \textbf{highlighted}. If the result of EpiS is the second best among all methods, it will be \underline{underlined}. }
\vspace{-0.3cm}

\centering
\resizebox{0.45\textwidth}{!}{%
\renewcommand{\arraystretch}{1.1}
\begin{tabular}{lrrrrrr} \hline 
Method  & Supervision & \multicolumn{1}{l}{$<1\uparrow$} & \multicolumn{1}{l}{$<2\uparrow$} & \multicolumn{1}{l}{$<4\uparrow$} & \multicolumn{1}{l}{Abs. $\downarrow$} & \multicolumn{1}{l}{Rel. $\downarrow$} \\ \hline
MVSNet \cite{yao2018mvsnet} & RGBD & 29.95 & 52.82 & 72.33 & 13.62  & 1.67 \\ 
VolRecon ~\cite{ren2023volrecon}& RGBD  & 44.22   & 65.62 & 80.19 & 7.87 & 1.00  \\
SparseNeuS \cite{long2022sparseneuss} & RGB  & 38.60 & 56.28  & 68.63  & 21.85 & 2.68 \\
\textbf{EpiS} & RGB  & \underline{43.97}  &  \textbf{66.16}  & \textbf{83.33} &  \textbf{7.79} & \textbf{0.99} \\ \hline
\hline   
\end{tabular}
}
\vspace{-0.6cm}

\label{table:dtudepth}
\end{table}

\subsubsection{Datasets}
We utilize the DTU dataset~\cite{aanaes2016large} for training, a multi-view stereo dataset comprising 124 scenes with ground truth point clouds and varying lighting conditions. Testing is conducted on the same 15 scenes as SparseNeuS, with the remaining scenes allocated for training. Depth maps rendered from the mesh serve as ground truth. Our framework, trained on the DTU dataset to ensure network generalization, employs 15 scenes for testing and the remaining 75 for training. Evaluation on testing scenes involves three views at 600 × 800 resolution, with each scene containing two sets of three images. Foreground masks provided by IDR are used for evaluation. To enhance memory efficiency during training, center-cropped images with 512 × 640 resolution are used. A simple threshold-based denoising strategy is applied to mitigate image noise. Additionally, we test our model on 7 challenging scenes from the BlendedMVS dataset~\cite{yao2020blendedmvs}, using one set of three images per scene with a resolution of 768 × 576. It's worth noting that in the per-scene fine-tuning stage, we optimize using the same three images without introducing new ones. Our experiments on both DTU and BlendedMVS datasets align with previous methods and include reporting Chamfer Distance for DTU and showcasing visual effects for BlendedMVS.

\subsubsection{Implementation}
Our model, developed using PyTorch~\cite{pytorch2018pytorch} and PyTorch Lightning, operates with an image resolution of $640 \times 512$ during training. Training occurs over 16 epochs, utilizing the Adam optimizer~\cite{kingma2014adam}, on a single 4090 GPU with a learning rate of $10^{-4}$. The batch size is set to 2, with 1024 rays sampled per batch. Both training and testing involve a hierarchical sampling strategy, initially sampling N coarse points uniformly on the ray and then employing importance sampling to sample additional N fine points on top of the coarse probability estimation, where $N coarse = 64$ and $N fine = 64$. The fine-level geometry encoding volumes are configured at a resolution of $96 \times 96 \times 96$. The sparse 3D CNN networks, structured akin to a U-Net, utilize a patch size of $5 \times 5$ for patch-based blending. During testing, the image resolution is adjusted to $800 \times 600$.

\begin{table*}[tb]
\caption{Quantitative results of \textbf{dense view} reconstruction on 15 testing scenes of DTU dataset. 
If the result of EpiS outperforms all other methods, it will be \textbf{highlighted}. If the result of EpiS is the second best among all methods, it will be \underline{underlined}.
}
\vspace{-0.5cm}

\resizebox{\textwidth}{!}{%
\renewcommand{\arraystretch}{1.1}
\begin{tabular}{lcccccccccccccccc} \\ \hline
Method          & 24   & 37   & 40   & 55   & 63   & 65   & 69   & 83   & 97   & 105  & 106  & 110  & 114  & 118  & 122  & \multicolumn{1}{l}{Mean$ \downarrow$} \\ \hline
NeRF~\cite{mildenhall2021nerf} & 1.90 & 1.60 & 1.85 & 0.58 & 2.28 & 1.27 & 1.47 & 1.67 & 2.05 & 1.07 & 0.88 & 2.53 & 1.06 & 1.15 & 0.96 & 1.49 \\
IDR~\cite{yariv2020multiview} & 1.63 & 1.87 & 0.63 & 0.48 & 1.04 & 0.79 & 0.77 & 1.33 & 1.16 & 0.76 & 0.67 & 0.90 & 0.42 & 0.51 & 0.53 & 0.90 \\
MVSDF~\cite{zhang2021learning} & 0.83 & 1.76 & 0.88 & 0.44 & 1.11 & 0.90 & 0.75 & 1.26 & 1.02 & 1.35 & 0.87 & 0.84 & 0.34 & 0.47 & 0.46 & 0.88 \\
VolSDF~\cite{yariv2021volume} & 1.14 & 1.26 & 0.81 & 0.49 & 1.25 & 0.70 & 0.72 & 1.29 & 1.18 & 0.70 & 0.66 & 1.08 & 0.42 & 0.61 & 0.55 & 0.86 \\
NeuS~\cite{wang2021neus} & 1.00 & 1.37 & 0.93 & 0.43 & 1.10 & 0.65 & 0.57 & 1.48 & 1.09 & 0.83 & 0.52 & 1.20 & 0.35 & 0.49 & 0.54 & 0.84 \\
Voxurf~\cite{wu2022voxurf} & 0.65 & 0.74 & 0.39 & 0.35 & 0.96 & 0.64 & 0.85 & 1.58 & 1.01 & 0.68 & 0.60 & 1.11 & 0.37 & 0.45 & 0.47 & 0.72 \\
COLMAP~\cite{schonberger2016structure} & 0.45 & 0.91 & 0.37 & 0.37 & 0.90 & 1.00 & 0.54 & 1.22 & 1.08 & 0.64 & 0.48 & 0.59 & 0.32 & 0.45 & 0.43 & 0.65 \\
\textbf{EpiS} & \underline{0.51} & \underline{0.81} & \underline{0.42} & \underline{0.44} & \textbf{0.82} & \underline{0.67} & \textbf{0.54} & \underline{1.24} & \textbf{0.95} & \underline{0.70} & \underline{0.54} & \underline{0.66} & \underline{0.38} & \textbf{0.44} & \underline{0.44} & \textbf{0.63}
  \\
\hline 
\end{tabular}
}
\vspace{-0.6cm}

\label{table:dtudense}
\end{table*}

\subsection{Evaluation Results}

\subsubsection{Sparse View Reconstruction on DTU}
In the DTU dataset~\cite{aanaes2016large}, we conduct sparse reconstruction using only 3 views, assessing performance through quantitative measures like Chamfer Distances between predicted meshes and ground truth point clouds, as shown in Table.~\ref{table:dtusparse} 
In contrast to MVS methods such as COLMAP~\cite{schonberger2016structure} and MVSNet~\cite{yao2018mvsnet}, our approach exhibits approximately a $11\%$ improvement over COLMAP, albeit slightly trailing behind MVSNet. It's worth noting that MVSNet requires ground truth depth maps and lacks the capability to render novel views.
Our method surpasses the state-of-the-art neural implicit reconstruction method SparseNeuS ~\cite{long2022sparseneuss}, achieving an $18\%$ enhancement in the generalizable experiment and a $16\%$ enhancement in the fine-tuning case. 
Moreover, our method outperforms VolRecon, which necessitates ground truth depth maps for training. 
Qualitative visualization in Figure.~\ref{fig:fig5} highlights our method's capacity to generate finer details and sharper boundaries than SparseNeuS.

\subsubsection{Depth map evaluation on DTU}
In this experiment, we evaluate depth estimation performance using SparseNeuS~\cite{long2022sparseneuss}a nd MVSNet~\cite{yao2018mvsnet} across all views in each scan. For each reference view, we select the top 4 source views based on view selection scores from~\cite{yao2018mvsnet} for depth rendering. The results in Table.~\ref{table:dtudepth} indicate that our method outperforms both MVSNet and SparseNeuS across all metrics, and even surpasses VolRecon. This finding aligns with the results reported in Table.~\ref{table:dtusparse}

\subsubsection{Full View Reconstruction on DTU}
For a comprehensive comparison, we also conduct a mesh reconstruction experiment using dense views (the first 49 views for all scenarios). As demonstrated in Table.~\ref{table:dtudense}, our method consistently outperforms previous per-scene optimization methods and MVS methods without the need for further training.

\subsubsection{Generalization on BlendedMVS}
\begin{figure}[tb]
    \centering
  \includegraphics[width=0.5\textwidth]{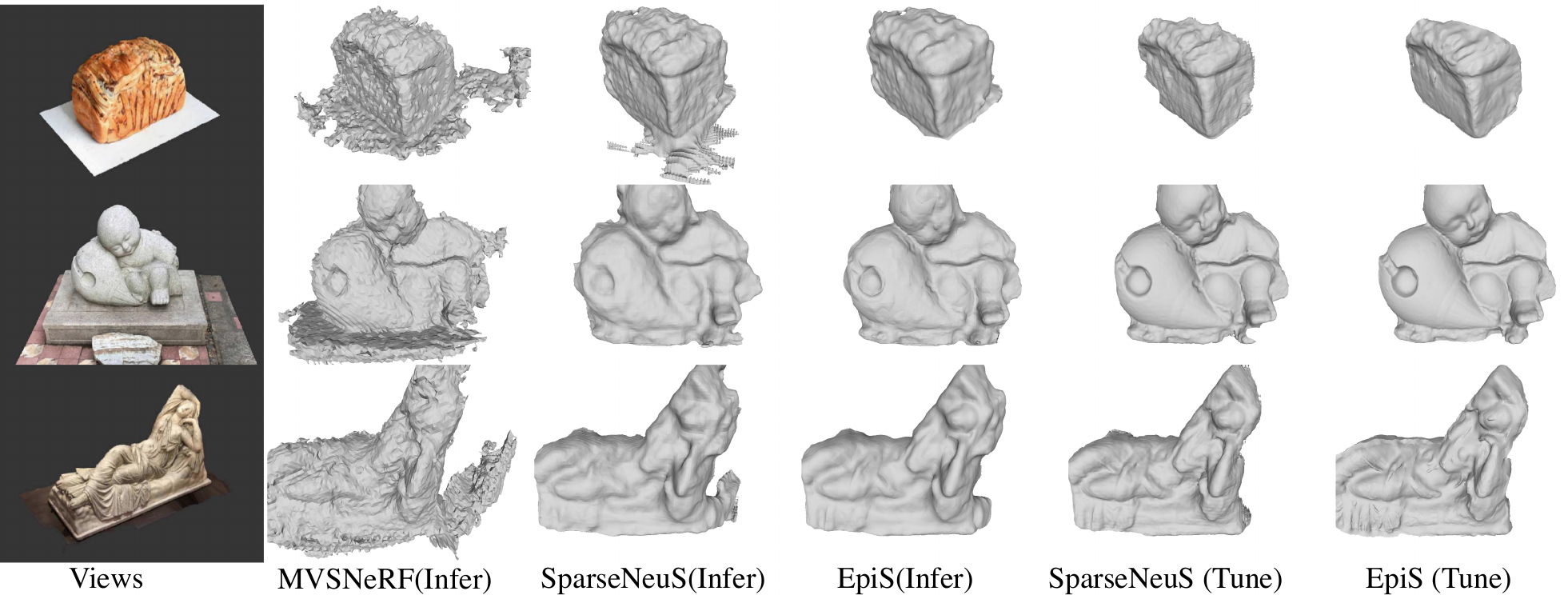}
    \caption{\textbf{Reconstruction results on the BlendedMVS dataset.} EpiS yields reasonably accurate estimation even without pre-training on BlendedMVS. Fine-tuning enhances EpiS's performance, leading to further improvements in accuracy.}
    \label{fig:fig7}
    \vspace{-0.5cm}
\end{figure}

To assess the generalization capability of our method, we extend our testing to BlendedMVS. The quantitative results are presented in Figure.~\ref{fig:fig7}. It is evident that EpiS exhibits superior performance in reconstructing details but may produce noisy boundaries compared to SparseNeuS.

\subsection{Ablation Studies}

\subsubsection{Epipolar Aggregation.}
In the first part of our ablation studies, we analyze the impact of removing the epipolar aggregation component from our framework. This adjustment results in the ray aggregation directly utilizing features from the cost-volume as input. The results in Table.~\ref{table:ablation} indicate that without epipolar aggregation, the ray aggregation process alone is insufficient to adequately capture the detailed geometry from the cost-volume data.

\subsubsection{Global Triplet Loss.}

Moving on to the second part, we examine the consequences of eliminating the global triplet loss function, which plays a crucial role in supervising the rendered depth. Without this loss function, our model fails to effectively leverage the relevant depth relationships provided by the pretrained large monocular depth model. As a result, there is a notable decline in performance.

\subsubsection{Local Gradient Loss.}
In the third part of our ablation studies, we focus on the removal of the local gradient loss function, which also contributes to supervising the rendered depth. Without this loss function, our model experiences a further deterioration in performance, underscoring the critical role it plays in maintaining the accuracy of our results.

\begin{table}[tb]
\centering
\caption{Ablation studies of epipolar transformer, global depth loss, and local depth loss on the DTU dataset. We report results on both sparise view setting and dense view setting. The best results will be \textbf{highlighted}. The second best results will be \underline{underlined}.}
\resizebox{0.45\textwidth}{!}{%
\renewcommand{\arraystretch}{1.1}
\begin{tabular}{l|c|ccccc|c} \hline
\multirow{2}{*}{Method}         & Sparse View Recon. & \multicolumn{5}{c|}{Depth Map Eval.} & Full View Recon. \\ 
 \cline{2-8}
& Chamfer$\downarrow$ & $<1\uparrow$ & $<2\uparrow$ &  $<4\uparrow$ & Abs.$\downarrow$ & Rel. $\downarrow$ & Chamfer$\downarrow$ \\
\hline
w/o Epipolar Aggregation & \underline{1.70} & \underline{31.04} & \underline{67.19} & \underline{82.55} & \underline{08.03} & \underline{1.10} & \underline{0.71}\\
w/o Global Triplet Loss & 1.86 & 20.64 & 38.86 & 56.48 & 15.49 & 2.11 & 0.83  \\ 
w/o Local Gradient Loss & 1.90 & 15.84 & 33.65 & 54.72 & 18.93 & 3.72 & 0.98  \\
\hline  
\textbf{EpiS} & \textbf{1.37} & \textbf{43.97}& \textbf{66.16}&\textbf{83.33}&\textbf{07.79}&\textbf{0.99}&\textbf{0.63}  \\
\hline  
\end{tabular}
}
\vspace{-0.4cm}
\label{table:ablation}
\end{table}

\begin{table}[tb]
\centering
\caption{Ablation study of the input view number on the DTU dataset. In this experiment, we modify the number of input views to test the performance of EpiS under different settings. The best results will be \textbf{highlighted}.The second best results will be \underline{underlined}.
}
\resizebox{0.45\textwidth}{!}{%
\begin{tabular}{c|c|cccc} \hline
Method & Supervision & 2 Views   & 3 Views   & 4 Views   & 6 Views   \\ \hline
VolRecon~\cite{ren2023volrecon} & RGBD  & 1.72 & 1.38 & 1.35 & 1.33 \\
\textbf{EpiS} & RGB  & \textbf{1.66} & \textbf{1.36} & \textbf{1.27} & \textbf{1.25} \\
 \hline 
\end{tabular}
}
\vspace{-0.4cm}
\label{table:numberofviews}
\end{table}

\subsubsection{Number of Input Views.}
In the final segment of our ablation studies, we investigate the performance of EpiSs with varying numbers of input views. All results are reported in Table.~\ref{table:numberofviews}. Our findings reveal that increasing the number of input views significantly enhances the performance of EpiS. This improvement is attributed to the increased effectiveness in eliminating occlusion, particularly evident in sparse view scenarios. The addition of more views effectively mitigates occlusion challenges, consequently leading to notable enhancements.

\section{Conclusion}
In this paper, we introduce EpiS, a generalizable method for sparse view neural surface reconstruction. Our approach incorporates an epipolar aggregation module to facilitate multi-view epipolar line fusion, guided by the cost volume. Additionally, we introduce a ray aggregation module to fuse information along the target ray. To regularize the depth learning process, we propose a global triplet loss and a local derivative loss, leveraging insights from a pretrained monocular depth model. Our model significantly outperforms state-of-the-art generalizable methods in neural surface reconstruction on the DTU dataset. Furthermore, the generalizability of our approach is demonstrated through compelling results on the BlendedMVS dataset.

{
    \small
    \bibliographystyle{ieeenat_fullname}
    \bibliography{main}

@String(CVPR= {IEEE Conf. Comput. Vis. Pattern Recog.})

@String(ECCV= {Eur. Conf. Comput. Vis.})

@String(TOG= {ACM Trans. Graph.})

@String(ICLR = {Int. Conf. Learn. Represent.})

@String(AAAI = {AAAI})

@String(CVPR  = {CVPR})

@String(ECCV  = {ECCV})

@String(TOG   = {ACM TOG})

@String(ICLR  = {ICLR})

@article{peng2024gens,
  title={GenS: Generalizable Neural Surface Reconstruction from Multi-View Images},
  author={Peng, Rui and Gu, Xiaodong and Tang, Luyang and Shen, Shihe and Yu, Fanqi and Wang, Ronggang},
  journal={Advances in Neural Information Processing Systems},
  volume={36},
  year={2024}
}

@inproceedings{wang2023neus2,
  title={Neus2: Fast learning of neural implicit surfaces for multi-view reconstruction},
  author={Wang, Yiming and Han, Qin and Habermann, Marc and Daniilidis, Kostas and Theobalt, Christian and Liu, Lingjie},
  booktitle={Proceedings of the IEEE/CVF International Conference on Computer Vision},
  pages={3295--3306},
  year={2023}
}

@inproceedings{zhou2022devnet,
  title={Devnet: Self-supervised monocular depth learning via density volume construction},
  author={Zhou, Kaichen and Hong, Lanqing and Chen, Changhao and Xu, Hang and Ye, Chaoqiang and Hu, Qingyong and Li, Zhenguo},
  booktitle={European Conference on Computer Vision},
  pages={125--142},
  year={2022},
  organization={Springer}
}

@inproceedings{wang2021ibrnet,
  title={Ibrnet: Learning multi-view image-based rendering},
  author={Wang, Qianqian and Wang, Zhicheng and Genova, Kyle and Srinivasan, Pratul P and Zhou, Howard and Barron, Jonathan T and Martin-Brualla, Ricardo and Snavely, Noah and Funkhouser, Thomas},
  booktitle={Proceedings of the IEEE/CVF Conference on Computer Vision and Pattern Recognition},
  pages={4690--4699},
  year={2021}
}

@article{mildenhall2021nerf,
  title={Nerf: Representing scenes as neural radiance fields for view synthesis},
  author={Mildenhall, Ben and Srinivasan, Pratul P and Tancik, Matthew and Barron, Jonathan T and Ramamoorthi, Ravi and Ng, Ren},
  journal={Communications of the ACM},
  volume={65},
  number={1},
  pages={99--106},
  year={2021},
  publisher={ACM New York, NY, USA}
}

@inproceedings{katharopoulos2020transformers,
  title={Transformers are rnns: Fast autoregressive transformers with linear attention},
  author={Katharopoulos, Angelos and Vyas, Apoorv and Pappas, Nikolaos and Fleuret, Fran{\c{c}}ois},
  booktitle={International conference on machine learning},
  pages={5156--5165},
  year={2020},
  organization={PMLR}
}

@article{bhat2023zoedepth,
  title={Zoedepth: Zero-shot transfer by combining relative and metric depth},
  author={Bhat, Shariq Farooq and Birkl, Reiner and Wofk, Diana and Wonka, Peter and M{\"u}ller, Matthias},
  journal={arXiv preprint arXiv:2302.12288},
  year={2023}
}

@article{ranftl2020towards,
  title={Towards robust monocular depth estimation: Mixing datasets for zero-shot cross-dataset transfer},
  author={Ranftl, Ren{\'e} and Lasinger, Katrin and Hafner, David and Schindler, Konrad and Koltun, Vladlen},
  journal={IEEE transactions on pattern analysis and machine intelligence},
  volume={44},
  number={3},
  pages={1623--1637},
  year={2020},
  publisher={IEEE}
}

@inproceedings{chen2021mvsnerf,
  title={Mvsnerf: Fast generalizable radiance field reconstruction from multi-view stereo},
  author={Chen, Anpei and Xu, Zexiang and Zhao, Fuqiang and Zhang, Xiaoshuai and Xiang, Fanbo and Yu, Jingyi and Su, Hao},
  booktitle={Proceedings of the IEEE/CVF International Conference on Computer Vision},
  pages={14124--14133},
  year={2021}
}

@article{aanaes2016large,
  title={Large-scale data for multiple-view stereopsis},
  author={Aan{\ae}s, Henrik and Jensen, Rasmus Ramsb{\o}l and Vogiatzis, George and Tola, Engin and Dahl, Anders Bjorholm},
  journal={International Journal of Computer Vision},
  volume={120},
  pages={153--168},
  year={2016},
  publisher={Springer}
}

@article{yao2020blendedmvs,
  title={BlendedMVS: A Large-scale Dataset for Generalized Multi-view Stereo Networks},
  author={Yao, Yao and Luo, Zixin and Li, Shiwei and Zhang, Jingyang and Ren, Yufan and Zhou, Lei and Fang, Tian and Quan, Long},
  journal={Computer Vision and Pattern Recognition (CVPR)},
  year={2020}
}

@inproceedings{yu2021pixelnerf,
  title={pixelnerf: Neural radiance fields from one or few images},
  author={Yu, Alex and Ye, Vickie and Tancik, Matthew and Kanazawa, Angjoo},
  booktitle={Proceedings of the IEEE/CVF Conference on Computer Vision and Pattern Recognition},
  pages={4578--4587},
  year={2021}
}

@article{yariv2020multiview,
title={Multiview Neural Surface Reconstruction by Disentangling Geometry and Appearance},
author={Yariv, Lior and Kasten, Yoni and Moran, Dror and Galun, Meirav and Atzmon, Matan and Ronen, Basri and Lipman, Yaron},
journal={Advances in Neural Information Processing Systems},
volume={33},
year={2020}
}

@article{yariv2021volume,
  title={Volume rendering of neural implicit surfaces},
  author={Yariv, Lior and Gu, Jiatao and Kasten, Yoni and Lipman, Yaron},
  journal={Advances in Neural Information Processing Systems},
  volume={34},
  pages={4805--4815},
  year={2021}
}

@inproceedings{oechsle2021unisurf,
  title={Unisurf: Unifying neural implicit surfaces and radiance fields for multi-view reconstruction},
  author={Oechsle, Michael and Peng, Songyou and Geiger, Andreas},
  booktitle={Proceedings of the IEEE/CVF International Conference on Computer Vision},
  pages={5589--5599},
  year={2021}
}

@inproceedings{schonberger2016structure,
  title={Structure-from-motion revisited},
  author={Schonberger, Johannes L and Frahm, Jan-Michael},
  booktitle={Proceedings of the IEEE conference on computer vision and pattern recognition},
  pages={4104--4113},
  year={2016}
}

@article{kingma2014adam,
  title={Adam: A method for stochastic optimization},
  author={Kingma, Diederik P and Ba, Jimmy},
  journal={arXiv preprint arXiv:1412.6980},
  year={2014}
}

@misc{pytorch2018pytorch,
  title={Pytorch},
  author={Pytorch, Automatic Differentiation In},
  year={2018}
}

@inproceedings{yao2018mvsnet,
  title={Mvsnet: Depth inference for unstructured multi-view stereo},
  author={Yao, Yao and Luo, Zixin and Li, Shiwei and Fang, Tian and Quan, Long},
  booktitle={Proceedings of the European conference on computer vision (ECCV)},
  pages={767--783},
  year={2018}
}

@inproceedings{gu2020cascade,
  title={Cascade cost volume for high-resolution multi-view stereo and stereo matching},
  author={Gu, Xiaodong and Fan, Zhiwen and Zhu, Siyu and Dai, Zuozhuo and Tan, Feitong and Tan, Ping},
  booktitle={Proceedings of the IEEE/CVF conference on computer vision and pattern recognition},
  pages={2495--2504},
  year={2020}
}

@inproceedings{ding2022transmvsnet,
  title={Transmvsnet: Global context-aware multi-view stereo network with transformers},
  author={Ding, Yikang and Yuan, Wentao and Zhu, Qingtian and Zhang, Haotian and Liu, Xiangyue and Wang, Yuanjiang and Liu, Xiao},
  booktitle={Proceedings of the IEEE/CVF Conference on Computer Vision and Pattern Recognition},
  pages={8585--8594},
  year={2022}
}

@inproceedings{middelberg2014scalable,
  title={Scalable 6-dof localization on mobile devices},
  author={Middelberg, Sven and Sattler, Torsten and Untzelmann, Ole and Kobbelt, Leif},
  booktitle={Computer Vision--ECCV 2014: 13th European Conference, Zurich, Switzerland, September 6-12, 2014, Proceedings, Part II 13},
  pages={268--283},
  year={2014},
  organization={Springer}
}

@inproceedings{zhou2021vmloc,
  title={Vmloc: Variational fusion for learning-based multimodal camera localization},
  author={Zhou, Kaichen and Chen, Changhao and Wang, Bing and Saputra, Muhamad Risqi U and Trigoni, Niki and Markham, Andrew},
  booktitle={Proceedings of the AAAI Conference on Artificial Intelligence},
  volume={35},
  number={7},
  pages={6165--6173},
  year={2021}
}

@article{wang2021neus,
  title={Neus: Learning neural implicit surfaces by volume rendering for multi-view reconstruction},
  author={Wang, Peng and Liu, Lingjie and Liu, Yuan and Theobalt, Christian and Komura, Taku and Wang, Wenping},
  journal={arXiv preprint arXiv:2106.10689},
  year={2021}
}

@inproceedings{ren2023volrecon,
  title={Volrecon: Volume rendering of signed ray distance functions for generalizable multi-view reconstruction},
  author={Ren, Yufan and Zhang, Tong and Pollefeys, Marc and S{\"u}sstrunk, Sabine and Wang, Fangjinhua},
  booktitle={Proceedings of the IEEE/CVF Conference on Computer Vision and Pattern Recognition},
  pages={16685--16695},
  year={2023}
}

@inproceedings{campbell2008using,
  title={Using multiple hypotheses to improve depth-maps for multi-view stereo},
  author={Campbell, Neill DF and Vogiatzis, George and Hern{\'a}ndez, Carlos and Cipolla, Roberto},
  booktitle={Computer Vision--ECCV 2008: 10th European Conference on Computer Vision, Marseille, France, October 12-18, 2008, Proceedings, Part I 10},
  pages={766--779},
  year={2008},
  organization={Springer}
}

@article{stereopsis2010accurate,
  title={Accurate, Dense, and Robust Multiview Stereopsis},
  author={Stereopsis, Robust Multiview},
  journal={IEEE TRANSACTIONS ON PATTERN ANALYSIS AND MACHINE INTELLIGENCE},
  volume={32},
  number={8},
  year={2010}
}

@inproceedings{ji2017surfacenet,
  title={Surfacenet: An end-to-end 3d neural network for multiview stereopsis},
  author={Ji, Mengqi and Gall, Juergen and Zheng, Haitian and Liu, Yebin and Fang, Lu},
  booktitle={Proceedings of the IEEE international conference on computer vision},
  pages={2307--2315},
  year={2017}
}

@article{ji2020surfacenet+,
  title={SurfaceNet+: An end-to-end 3D neural network for very sparse multi-view stereopsis},
  author={Ji, Mengqi and Zhang, Jinzhi and Dai, Qionghai and Fang, Lu},
  journal={IEEE Transactions on Pattern Analysis and Machine Intelligence},
  volume={43},
  number={11},
  pages={4078--4093},
  year={2020},
  publisher={IEEE}
}

@article{kar2017learning,
  title={Learning a multi-view stereo machine},
  author={Kar, Abhishek and H{\"a}ne, Christian and Malik, Jitendra},
  journal={Advances in neural information processing systems},
  volume={30},
  year={2017}
}

@article{kutulakos2000theory,
  title={A theory of shape by space carving},
  author={Kutulakos, Kiriakos N and Seitz, Steven M},
  journal={International journal of computer vision},
  volume={38},
  pages={199--218},
  year={2000},
  publisher={Springer}
}

@article{gropp2020implicit,
  title={Implicit geometric regularization for learning shapes},
  author={Gropp, Amos and Yariv, Lior and Haim, Niv and Atzmon, Matan and Lipman, Yaron},
  journal={arXiv preprint arXiv:2002.10099},
  year={2020}
}

@article{lhuillier2005quasi,
  title={A quasi-dense approach to surface reconstruction from uncalibrated images},
  author={Lhuillier, Maxime and Quan, Long},
  journal={IEEE transactions on pattern analysis and machine intelligence},
  volume={27},
  number={3},
  pages={418--433},
  year={2005},
  publisher={IEEE}
}

@inproceedings{schonberger2016pixelwise,
  title={Pixelwise view selection for unstructured multi-view stereo},
  author={Sch{\"o}nberger, Johannes L and Zheng, Enliang and Frahm, Jan-Michael and Pollefeys, Marc},
  booktitle={Computer Vision--ECCV 2016: 14th European Conference, Amsterdam, The Netherlands, October 11-14, 2016, Proceedings, Part III 14},
  pages={501--518},
  year={2016},
  organization={Springer}
}

@article{tola2012efficient,
  title={Efficient large-scale multi-view stereo for ultra high-resolution image sets},
  author={Tola, Engin and Strecha, Christoph and Fua, Pascal},
  journal={Machine Vision and Applications},
  volume={23},
  pages={903--920},
  year={2012},
  publisher={Springer}
}

@article{liang2024retr,
  title={ReTR: Modeling Rendering Via Transformer for Generalizable Neural Surface Reconstruction},
  author={Liang, Yixun and He, Hao and Chen, Yingcong},
  journal={Advances in Neural Information Processing Systems},
  volume={36},
  year={2024}
}

@inproceedings{genova2019learning,
  title={Learning shape templates with structured implicit functions},
  author={Genova, Kyle and Cole, Forrester and Vlasic, Daniel and Sarna, Aaron and Freeman, William T and Funkhouser, Thomas},
  booktitle={Proceedings of the IEEE/CVF International Conference on Computer Vision},
  pages={7154--7164},
  year={2019}
}

@inproceedings{mescheder2019occupancy,
  title={Occupancy networks: Learning 3d reconstruction in function space},
  author={Mescheder, Lars and Oechsle, Michael and Niemeyer, Michael and Nowozin, Sebastian and Geiger, Andreas},
  booktitle={Proceedings of the IEEE/CVF conference on computer vision and pattern recognition},
  pages={4460--4470},
  year={2019}
}

@inproceedings{michalkiewicz2019implicit,
  title={Implicit surface representations as layers in neural networks},
  author={Michalkiewicz, Mateusz and Pontes, Jhony K and Jack, Dominic and Baktashmotlagh, Mahsa and Eriksson, Anders},
  booktitle={Proceedings of the IEEE/CVF International Conference on Computer Vision},
  pages={4743--4752},
  year={2019}
}

@inproceedings{niemeyer2019occupancy,
  title={Occupancy flow: 4d reconstruction by learning particle dynamics},
  author={Niemeyer, Michael and Mescheder, Lars and Oechsle, Michael and Geiger, Andreas},
  booktitle={Proceedings of the IEEE/CVF international conference on computer vision},
  pages={5379--5389},
  year={2019}
}

@inproceedings{park2019deepsdf,
  title={Deepsdf: Learning continuous signed distance functions for shape representation},
  author={Park, Jeong Joon and Florence, Peter and Straub, Julian and Newcombe, Richard and Lovegrove, Steven},
  booktitle={Proceedings of the IEEE/CVF conference on computer vision and pattern recognition},
  pages={165--174},
  year={2019}
}

@inproceedings{peng2020convolutional,
  title={Convolutional occupancy networks},
  author={Peng, Songyou and Niemeyer, Michael and Mescheder, Lars and Pollefeys, Marc and Geiger, Andreas},
  booktitle={Computer Vision--ECCV 2020: 16th European Conference, Glasgow, UK, August 23--28, 2020, Proceedings, Part III 16},
  pages={523--540},
  year={2020},
  organization={Springer}
}

@article{liu2020neural,
  title={Neural sparse voxel fields},
  author={Liu, Lingjie and Gu, Jiatao and Zaw Lin, Kyaw and Chua, Tat-Seng and Theobalt, Christian},
  journal={Advances in Neural Information Processing Systems},
  volume={33},
  pages={15651--15663},
  year={2020}
}

@inproceedings{liu2020dist,
  title={Dist: Rendering deep implicit signed distance function with differentiable sphere tracing},
  author={Liu, Shaohui and Zhang, Yinda and Peng, Songyou and Shi, Boxin and Pollefeys, Marc and Cui, Zhaopeng},
  booktitle={Proceedings of the IEEE/CVF Conference on Computer Vision and Pattern Recognition},
  pages={2019--2028},
  year={2020}
}

@article{muller2022instant,
  title={Instant neural graphics primitives with a multiresolution hash encoding},
  author={M{\"u}ller, Thomas and Evans, Alex and Schied, Christoph and Keller, Alexander},
  journal={ACM Transactions on Graphics (ToG)},
  volume={41},
  number={4},
  pages={1--15},
  year={2022},
  publisher={ACM New York, NY, USA}
}

@inproceedings{oechsle2019texture,
  title={Texture fields: Learning texture representations in function space},
  author={Oechsle, Michael and Mescheder, Lars and Niemeyer, Michael and Strauss, Thilo and Geiger, Andreas},
  booktitle={Proceedings of the IEEE/CVF International Conference on Computer Vision},
  pages={4531--4540},
  year={2019}
}

@inproceedings{pumarola2021d,
  title={D-nerf: Neural radiance fields for dynamic scenes},
  author={Pumarola, Albert and Corona, Enric and Pons-Moll, Gerard and Moreno-Noguer, Francesc},
  booktitle={Proceedings of the IEEE/CVF Conference on Computer Vision and Pattern Recognition},
  pages={10318--10327},
  year={2021}
}

@inproceedings{sun2022direct,
  title={Direct voxel grid optimization: Super-fast convergence for radiance fields reconstruction},
  author={Sun, Cheng and Sun, Min and Chen, Hwann-Tzong},
  booktitle={Proceedings of the IEEE/CVF Conference on Computer Vision and Pattern Recognition},
  pages={5459--5469},
  year={2022}
}

@inproceedings{niemeyer2020differentiable,
  title={Differentiable volumetric rendering: Learning implicit 3d representations without 3d supervision},
  author={Niemeyer, Michael and Mescheder, Lars and Oechsle, Michael and Geiger, Andreas},
  booktitle={Proceedings of the IEEE/CVF Conference on Computer Vision and Pattern Recognition},
  pages={3504--3515},
  year={2020}
}

@inproceedings{zhang2021learning,
  title={Learning signed distance field for multi-view surface reconstruction},
  author={Zhang, Jingyang and Yao, Yao and Quan, Long},
  booktitle={Proceedings of the IEEE/CVF International Conference on Computer Vision},
  pages={6525--6534},
  year={2021}
}

@article{yu2022monosdf,
  title={Monosdf: Exploring monocular geometric cues for neural implicit surface reconstruction},
  author={Yu, Zehao and Peng, Songyou and Niemeyer, Michael and Sattler, Torsten and Geiger, Andreas},
  journal={Advances in neural information processing systems},
  volume={35},
  pages={25018--25032},
  year={2022}
}

@inproceedings{johari2022geonerf,
  title={Geonerf: Generalizing nerf with geometry priors},
  author={Johari, Mohammad Mahdi and Lepoittevin, Yann and Fleuret, Fran{\c{c}}ois},
  booktitle={Proceedings of the IEEE/CVF Conference on Computer Vision and Pattern Recognition},
  pages={18365--18375},
  year={2022}
}

@article{zhou2024dynpoint,
  title={Dynpoint: Dynamic neural point for view synthesis},
  author={Zhou, Kaichen and Zhong, Jia-Xing and Shin, Sangyun and Lu, Kai and Yang, Yiyuan and Markham, Andrew and Trigoni, Niki},
  journal={Advances in Neural Information Processing Systems},
  volume={36},
  year={2024}
}

@inproceedings{long2022sparseneuss,
  title={Sparseneus: Fast generalizable neural surface reconstruction from sparse views},
  author={Long, Xiaoxiao and Lin, Cheng and Wang, Peng and Komura, Taku and Wang, Wenping},
  booktitle={European Conference on Computer Vision},
  pages={210--227},
  year={2022},
  organization={Springer}
}

@inproceedings{wu2022voxurf,
    title={Voxurf: Voxel-based Efficient and Accurate Neural Surface Reconstruction},
    author={Tong Wu and Jiaqi Wang and Xingang Pan and Xudong Xu and Christian Theobalt and Ziwei Liu and Dahua Lin},
    booktitle={International Conference on Learning Representations (ICLR)},
    year={2023},
}

@inproceedings{wang2024dust3r,
  title={{DUSt3R}: Geometric {3D} Vision Made Easy},
  author={Wang, Shuzhe and Leroy, Vincent and Cabon, Yohann and Chidlovskii, Boris and Revaud, Jerome},
  booktitle={Proceedings of the IEEE/CVF Conference on Computer Vision and Pattern Recognition},
  pages={20697--20709},
  year={2024}
}

@inproceedings{wang2025vggt,
  title={{VGGT}: Visual geometry grounded transformer},
  author={Wang, Jianyuan and Chen, Minghao and Karaev, Nikita and Vedaldi, Andrea and Rupprecht, Christian and Novotny, David},
  booktitle={Proceedings of the IEEE/CVF Conference on Computer Vision and Pattern Recognition},
  pages={5294--5306},
  year={2025}
}

@inproceedings{zhou2026page,
  title={Page-4d: Disentangled pose and geometry estimation for vggt-4d perception},
  author={Zhou, Kaichen and Wang, Yuhan and Chen, Grace and Beaudouin, Gaspard and Zhan, Fangneng and Liang, Paul and Wang, Mengyu},
  booktitle={International Conference on Learning Representations},
  volume={2026},
  pages={36401--36414},
  year={2026}
}
}


\end{document}